\documentclass{article}
\usepackage{ijcai20}

\usepackage{times}
\usepackage{soul}
\usepackage{url}
\usepackage[hidelinks]{hyperref}
\usepackage[utf8]{inputenc}
\usepackage[small]{caption}
\usepackage{graphicx}
\usepackage{amsmath}
\usepackage{amsthm}
\usepackage{booktabs}
\usepackage{algorithm}
\usepackage{algorithmic}
\usepackage{amsfonts}
\usepackage{subfigure} 
\title{KoGuN: Accelerating Deep Reinforcement Learning via Integrating Human Suboptimal Knowledge}

\author{
Peng Zhang$^{1,2}$
\and
Jianye Hao$^{1,2,3,}$\footnote{Corresponding author.}\and
Weixun Wang$^1$\and
Hongyao Tang$^1$\and
Yi Ma$^1$\and
\\Yihai Duan$^1$\And
Yan Zheng$^1$
\affiliations
$^1$College of Intelligence and Computing, Tianjin University
$^2$Noah's Ark Lab, Huawei
$^3$Tianjin Key Lab of Machine Learning
\emails
\{zhangpeng123, jianye.hao, wxwang, bluecontra, mayi, duanyihai, yanzheng\}@tju.edu.cn
}

\begin{document}

\maketitle

\begin{abstract}  
Reinforcement learning agents usually learn from scratch, which requires a large number of interactions with the environment.
This is quite different from the learning process of human. When faced with a new task, human naturally have the common sense and use the prior knowledge to derive an initial policy and guide the learning process afterwards. Although the prior knowledge may be not fully applicable to the new task, the learning process is significantly sped up since the initial policy ensures a quick-start of learning and intermediate guidance allows to avoid unnecessary exploration. Taking this inspiration, we propose knowledge guided policy network (KoGuN), a novel framework that combines human prior suboptimal knowledge with reinforcement learning. Our framework consists of a fuzzy rule controller to represent human knowledge and a refine module to fine-tune suboptimal prior knowledge. The proposed framework is end-to-end and can be combined with existing policy-based reinforcement learning algorithm. We conduct experiments on both discrete and continuous control tasks. The empirical results show that our approach, which combines human suboptimal knowledge and RL, achieves significant improvement on learning efficiency of flat RL algorithms, even with very low-performance human prior knowledge. 
\end{abstract}

\section{Introduction} \label{intro}
Deep reinforcement learning (DRL) algorithms have been applied in a range of challenging domains, from playing games \cite{dqn,mastergo} to robotic control \cite{robotic}. The combination of RL and high-capacity function approximators such as neural networks holds the promise of automating a wide range of decision making and control tasks, but application of these methods in real-world domains has been hampered by the challenge of sample complexity. Even relatively simple tasks can require millions of steps of data collection, and complex behaviors with high-dimensional observations might need substantially more.

In human's learning process, they rarely learn to master a new task from scratch. In contrast, human naturally leverage knowledge obtained in similar tasks to derive an rough policy and guide the learning process afterwards. Although these knowledge may be not completely compatible with the new task, human can adjust the policy in the following learning process. As a result, human can learn much faster than RL agents. Therefore, integrating human knowledge into reinforcement learning algorithms is promising to boost the learning process. 

Combining human knowledge has been studied before in supervised learning \cite{augfeat,dl2,nss,harnessing}. 
An important line of works that leverage human knowledge in sequential decision-making problem is imitation learning \cite{dagger,gail,bc,dqfd}. Imitation learning leverages human knowledge by learning from expert trajectories that collected in the same task that we aim to solve. 
Demonstration data is the concrete instance of human knowledge under a certain task and can be seen as low-level representation of human knowledge.
We expect to leverage high-level knowledge (e.g., common sense) to assist learning under unseen tasks (thus no demonstration data). Moreover, high-level knowledge can be easily shared with other similar tasks.

To leverage human knowledge, a major challenge is obtaining the representation of the provided knowledge. Under most circumstances, the provided knowledge is imprecise and uncertain, and even cover only a small part of the state space. As a consequence, the conventional approaches such as classical bivalent logic rules do not provide an adequate model for modes of reasoning which are approximate rather than exact \cite{fuzzysets,krfl}. 
In contrast, Fuzzy Logic, which may be viewed as an extension of classical logical systems, provides an effective conceptual framework for dealing with the problem of knowledge representation in an environment of uncertainty and imprecision. 
In this paper, we propose a novel knowledge guided policy network (KoGuN), which can integrate human knowledge into RL algorithms in an end-to-end manner. The proposed policy framework consists of a trainable knowledge controller and a refine module. On one hand, the knowledge controller contains a set of fuzzy logic rules to represent human knowledge which can be continuously optimized together with the whole policy; on the other hand, the refine module undertakes the role of refining the provided suboptimal and imprecise knowledge. Finally, the combination of the knowledge controller and the refine module helps "warm-start" the learning process and enables the agent to learn faster.
Generally, the proposed framework can be combined with existing policy-based reinforcement learning algorithms to accelerate their learning processes. We evaluate our method on discrete and continuous control tasks and the experimental results show that our approach achieves significant improvement on learning efficiency of RL algorithms.

The remainder of this paper is organized as follows:  Section~\ref{background} introduces deep reinforcement learning and fuzzy logic. 
Section~\ref{method} demonstrates the overall proposed policy framework. Section~\ref{exp} gives experimental results on both discrete and continuous tasks and demonstrates the effectiveness of the proposed policy framework.
Finally, concluding remarks are provided in Section~\ref{conc}.

\begin{figure*}[t]
\centering
\includegraphics[width=0.8\textwidth]{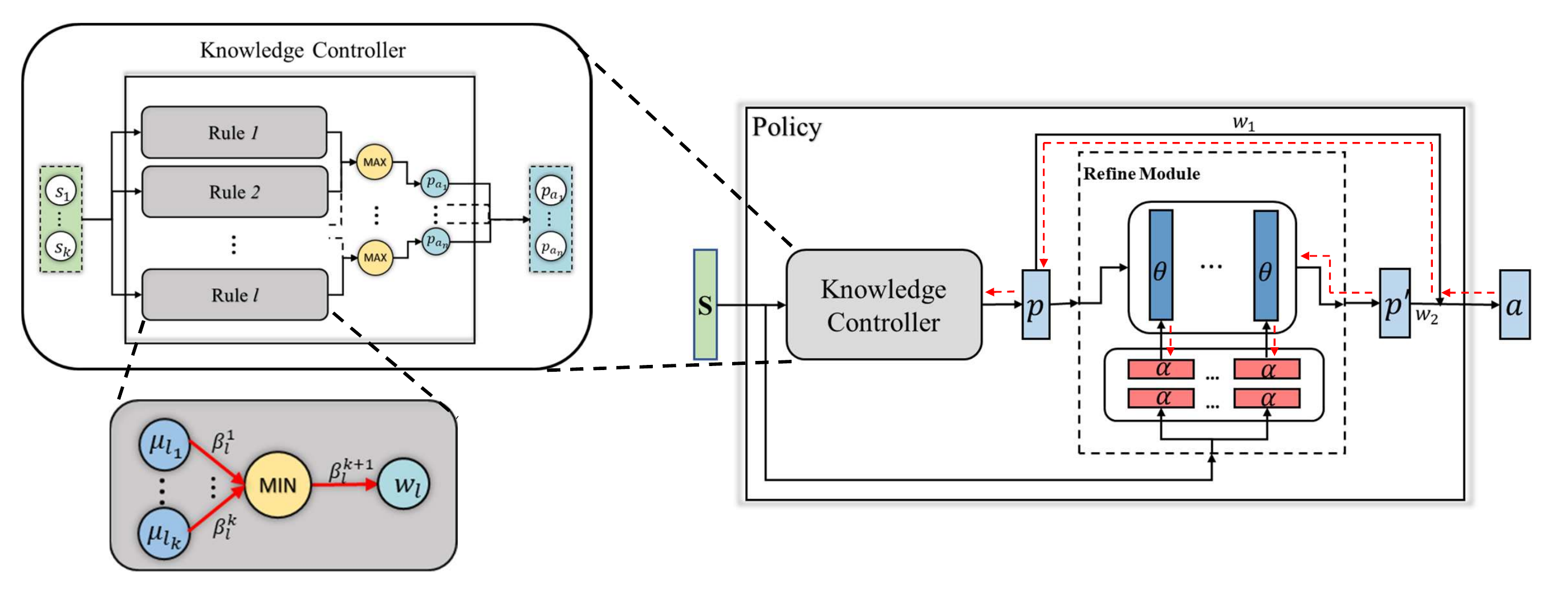} 
\caption{The overall architecture of knowledge guided policy network. It consists of a knowledge controller for human knowledge representation and a refine module ($dashed ~box$) for policy refinement. $red ~dashed ~lines$ indicate the gradient flow.}
\label{kgpn}
\end{figure*}

\section{Preliminary} \label{background}

\subsection{Deep Reinforcement Learning}
An Markov decision process (MDP) can be defined by a 5-tuple $(\mathcal{S}, \mathcal{A}, \mathcal{P}, \mathcal{R}, \gamma)$, where $\mathcal{S}$ is the state space, $\mathcal{A}$ is the action space, $\mathcal{P}$ is the transition function, $\mathcal{R}$ is the reward function and $\gamma$ is the discount factor \cite{suttonbook}. An RL agent interacts with the MDP following a policy $\pi\left(a_{t} | s_{t}\right)$, which is a mapping of state space to action space. The agent aims to learn a policy that maximize the expected discounted total reward. The state value function $V_{\pi}(s)$ is an estimate of the expected future reward that can be obtained from state $s$ when following policy $\pi$:
\begin{equation}
    V_{\pi}(s)=\mathbb{E}_\pi\left[\sum_{k=0}^{\infty}\gamma^{k}r_{t+k+1}|S_t=s\right]
\end{equation}

Several effective policy-based algorithms have been proposed in the literature. TRPO \cite{robotic} and ACKTR \cite{acktr} both update the policy subject a constraint in the state-action space (trust region). Proximal policy optimization (PPO) \cite{ppo} designs a clipped surrogate objective that approximates the regularization. PPO aims to maximize the following objective function:

\begin{align}
    J(\theta)= \mathbb{E}_{(s_t,a_t)\sim\pi_{\theta_{old}}}\left[min\left(\frac{\pi_{\theta}(a_t|s_t)}{\pi_{\theta_{old}}(a_t|s_t)}\hat{A}_t(s_t,a_t), 
    \notag\right.\right.
    \phantom{=\;\;}
    \\ 
    \left.\left. clip\left(\frac{\pi_{\theta}(a_t|s_t)}{\pi_{\theta_{old}}(a_t|s_t)},1-\epsilon,1+\epsilon\right)\hat{A}_t(s_t,a_t)\right)\right]
\end{align}
Here $\hat{A}_t(s_t,a_t)$ is an estimator of the advantage function at timestep $t$ and in this paper we use generalized advantage estimation (GAE) \cite{gae} to calculate the advantage function. 
When dealing with tasks with discrete action space, the policy network usually outputs the action distribution by applying $softmax$ on the logits. For continuous action space, the policy network normally outputs mean and standard deviation of a Gaussian distribution.

\subsection{Fuzzy Logic and Fuzzy Rules} \label{flfr}
Fuzzy logic is based on fuzzy set theory \cite{fuzzysets}. A fuzzy set is an extension of the classical notion of crisp set. Crisp sets allow only full membership or no membership at all. Compared with crisp sets, fuzzy sets allow partial membership. An element may partially belong to a fuzzy set. The membership of an element $x$ in a crisp set $S$ can be described by a membership function $\mu_S(x)$, where:
\begin{equation}
    \mu_S(x)=
    \left\{
        \begin{array}{lr}
            1 & x \in S \\
            0 & x \notin S
        \end{array}
    \right.
\end{equation}
But for a fuzzy set $F$, the membership of $x$ in it can be described by a membership function $\mu_F(x)$ with range from 0 to 1:
\begin{equation}
    \mu_F:X\to[0, 1]
\end{equation}
where $X$ refers to the universal set defined in a specific problem. 

Operations in classical crisp set theory can also be extended to fuzzy set theory. Assuming that $F_1$ and $F_2$ are two fuzzy sets. $\mu_{F_1}$ and $\mu_{F_2}$ are membership functions of $F_1$ and $F_2$ respectively. The the $union$ of $F_1$ and $F_2$ is a fuzzy set whose membership function is:
\begin{equation}
    \mu_{F_1 \cup F_2}(x) = max[\mu_{F_1}(x), \mu_{F_2}(x)]
\end{equation}
The $intersection$ of $F_1$ and $F_2$ is the fuzzy set with membership function:
\begin{equation}
    \mu_{F_1 \cap F_2}(x) = min[\mu_{F_1}(x), \mu_{F_2}(x)]
\end{equation}

A fuzzy rule is usually in the form of `IF $X$ is $A$ and $Y$ is $B$ THEN $Z$ is $C$'. `$X$ is $A$' and `$Y$ is $B$' are called preconditions of the fuzzy rule and `$Z$ is $C$' is called conclusion of the rule. $X$, $Y$ and $Z$ are variables. $A$, $B$ and $C$ are fuzzy sets which are normally called as linguistic values. A fuzzy rule takes the observation values of $X$ and $Y$ as input and outputs the value of $Z$. To get the reasoning conclusion of a fuzzy rule, we first calculate the truth value $T$ of each precondition. Then applying conjunction operator to these truth values, we finally get the conclusion's strength $w$ ($w$ can also be seen as the satisfaction level of the rule):
\begin{equation}
 w  = min(T_1, T_2)      
    = min[\mu_A(x_0), \mu_B(y_0)]      
\end{equation}
Here $x_0$ and $y_0$ are observation values for $X$ and $Y$. $T_1$ and $T_2$ are truth values for preconditions `$X$ is $A$' and `$Y$ is $B$'. $minimum$ operator is used here as conjunction operator. Finally, to get the value of $Z$, we need to match the conclusion's strength on the domain of $Z$. One common used matching method is applying the inverse function of the membership function of `$C$' on the strength:
\begin{equation}
    z=\mu_C^{-1}(w)
\end{equation}
Note that inverse functions can be defined only for monotonic functions, and in this paper we use simple monotonic linear functions as membership functions of conclusions. Other matching methods can be found in \cite{matchmethod}.

\subsection{Knowledge Representation Using Fuzzy Rules} \label{kr}
Human common sense knowledge is often imprecise and uncertain. Conventional knowledge representation approaches such as hard rules, which are based on classical bivalent logic, are inappropriate to represent this type of knowledge \cite{krfl}. Fuzzy logic was motivated in large measure by the need for a conceptual framework which can address the issues of uncertainty and lexical imprecision and thus are suitable for representing human imprecise knowledge.

To illustrate how human knowledge could be translated into fuzzy rules, consider an example of learning to drive a car. When the learner sits in the driver's seat and observes that the car is off the road to the right, he would follow the common sense and turn the steering wheel to the left to keep the car in the middle of the road. 
Nevertheless, he is unaware of how many degrees precisely should the steering wheel be rotated. The only thing he is sure is that the farther the car is off the middle of road, the larger angle of the steering wheel should be rotated. Alongside the learning process continues, the learner gradually learns an optimal policy on how to drive a car. The prior knowledge of keeping the car in the middle of road can be expressed in the form of `IF...THEN...', for instance, `IF the car's deviation from the middle of road to the right is large, THEN the steering wheel should be turned to the left'. But we do not have an accurate definition of `large'. Therefore, we can use a fuzzy set to define `large'. Hence, the common sense mentioned before can be translated into a fuzzy rule: `IF $D_r$ is $large$ THEN $Action$ is $left$'. Here $D_r$ represents the car's deviation distance from the middle of road to the right. $large$ is a fuzzy set, and its membership function can be defined as a simple linear function, for example, $y=clip[(0.5x-1), 0, 1]$. Such a membership function means that the larger the deviation value $x$ is, the greater the membership $y$ is, resulting in a greater probability of turning the steering wheel to the left.

\section{Knowledge Guided Policy Network} \label{method}
In this section, we propose a novel end-to-end framework to leverage human knowledge where the priors are continuously optimized with the whole policy. The proposed policy framework is called knowledge guided policy network (KoGuN) and the overall architecture of KoGuN is shown in Figure~\ref{kgpn}. KoGuN consists of a trainable knowledge controller and a refine module.
In Section~\ref{fc}, we describe the architecture of the knowledge controller which undertakes the role of incorporating human knowledge into the policy framework in an end-to-end manner. In Section~\ref{refine}, we introduce the architecture of the refine module, which fine-tunes the prior knowledge since the rule knowledge is normally suboptimal and even cover only a small part of the state space. 
Finally in Section~\ref{combine}, we demonstrate how to combine the refine module and the knowledge controller, forming a complete policy framework.

\subsection{Knowledge Controller} \label{fc}
Knowledge controller module plays the role of knowledge representation. However, there is no guarantee that provided human knowledge can fit perfectly into the current task, leading to knowledge mismatch problem. The proposed knowledge controller alleviates the problem by introducing trainable weights.

As shown in Figure~\ref{kgpn} ($top$ $left$), the knowledge controller $\phi(s)$ , containing a few fuzzy rules translated from knowledge provided by human, takes state information as input and outputs an action preference vector $\boldsymbol{p}$, which represents the tendency of the controller in a state. Each rule corresponds to one action and has the form of:
\begin{itemize}
\item $Rule$ $l$: IF $S_1$ is $M_{l_1}$ and $S_2$ is $M_{l_2}$ and ... and $S_k$ is $M_{l_k}$ THEN $Action$ is $a_j$
\end{itemize}
Here $S_i$ are variables that describe different parts of system states. $M_{l_i}$ are fuzzy sets corresponding to $S_i$. The conclusion of the rule indicates the corresponding action $a_j$. Now assume that we have observation values $s_1$...$s_k$ for $S_1$...$S_k$, we can get truth values for each precondition by applying membership functions to the observed values: $\mu_{l_i}(s_i)$, where $\mu_{l_i}$ are membership functions of the fuzzy set $M_{l_i}$. Hence, the strength of $Rule$ $l$ can be calculated as described in Section~\ref{flfr}.
To mitigate the knowledge mismatch problem, inspired by \cite{rlfuzzy}, we add trainable weights $\beta$ to each rule, which enable the knowledge controller to learn to adapt to current task. For each rule $l$ there are $k+1$ weights corresponding to it, here $k$ is the number of the preconditions. $\beta_{l}^1...\beta_{l}^k$ are assigned to each precondition and $\beta_{l}^{k+1}$ is assigned to the entire rule. The adjustment of the weights $\beta_{l}^1...\beta_{l}^k$ entails the adjustment of corresponding membership functions and the weight $\beta_{l}^{k+1}$ implies the importance of the rule. 
With trainable weights, the knowledge controller can be optimized like a neural network. The trainable weights are labelled as red lines in Figure~\ref{kgpn} ($bottom$ $left$). Finally, the strength of $Rule$ $l$ can be calculated by:
\begin{equation}
    w_l=\beta_{l}^{k+1}min[\beta_{l}^1\mu_{l_1}(s_1),\beta_{l}^2\mu_{l_2}(s_2),...,\beta_{l}^k\mu_{l_k}(s_k)]
\end{equation}
In this paper, these trainable weights are initialized to 1 at beginning for not disturbing the prior knowledge. Note that the trainable weights can also be set according to prior knowledge. For example, if the confidence corresponding to one of the provided rules is high, the rule then can be assigned a higher weight.

For discrete action space, the rule strength can be seen as this rule's preference for the corresponding action. Different rules cover different parts of the state space. Thus the relation of different rules for the same action is $or$ and the preference $p_{a_i}$ for action $a_i$ can be calculated by using $maximum$ operator to take the largest strength value of those rules corresponding to action $a_i$. Finally, we get the action preference vector $\boldsymbol{p}$:
\begin{equation} \label{preference}
\boldsymbol{p} = \phi_{\beta}(s) = [p_{a_1}, p_{a_2}, ...,p_{a_{|\mathcal{A}|}}]
\end{equation}
The vector $\boldsymbol{p}$ represents all the rules' preference for actions in a state. 

For continuous action space, we need to further map the rule strength to a continuous action value. As described in Section~\ref{flfr}, one common used method is applying the inverse membership function of the conclusion. As an example, consider an $n$-dimensional continuous action space. In this paper, one rule is designed only for one of the $n$ dimensions and we assume that Rule $l$ is designed for the $d$th dimension. The reasoning result of Rule $l$ can be calculated by:
\begin{equation} 
p_{d} = \mu_{a_j}^{-1}(w_l)
\end{equation}
 If there are more than one rules for the same action dimension, the weighted average of these rules' action value weighted by the rule strength is used as the final output of these rules. Finally, we get the action preference vector $\boldsymbol{p}$:
 \begin{equation} 
\boldsymbol{p} = \phi_{\beta}(s) = [p_{1}, p_{2}, ...,p_{d},...,p_{n}]
\end{equation}

\subsection{Knowledge Refine Module}\label{refine}
The knowledge controller, representing the prior knowledge provided by human, is only a very rough policy that may not cover the whole state space. To obtain an optimal or near optimal policy, the knowledge controller needs to be extended and further refined. We import a refine module $g(\cdot)$ to refine and complete the rough rule-based policy. The refine module $g(\cdot)$ should take the preference vector $\boldsymbol{p}$ as input and output the refined action preference vector $\boldsymbol{p'}$. The refined process could be regarded as correction and complement of the rough rule-based policy, which may only cover parts of the whole state space. How to refine $\boldsymbol{p}$ should be based on current state $s$. Here we propose two alternative ways to approximate the refine module $g(\cdot)$. The most straightforward idea is to approximate $g(\cdot)$ using a neural network. i.e., concatenating state and preference vector as the input of the neural network:

\begin{equation} \label{fc_rm}
\boldsymbol{p'} = g_{\theta}(s,\boldsymbol{p})
\end{equation}
Here $\theta$ is the parameters of the neural network.

However, the refinement of the knowledge controller can be different in different states, which means that the mapping from the action preference vector $\boldsymbol{p}$ to the refined preference vector $\boldsymbol{p'}$ can be different during the change of states. This requires the refine module $g_{\theta}(\cdot)$ to change drastically as the state $s$ changes. The parameters $\theta$ is required to capture such complex relation. One previous work \cite{fastweights} shows that using a learning feed-forward network to generate weights for a second network is suitable for this kind of context-dependent function. Thus we proposed that approximating the refine module $g(\cdot)$ using hypernetworks \cite{hypernet}, which leverage the idea of \cite{fastweights}. Moreover, using a feed-forward network to generate weights for another netowrk, hypernetworks are more in line with the semantic of the refine module. The first network takes state as input and generates weights for the second one. The second network takes action preference vector $\boldsymbol{p}$ as input and refines it, finally outputs the refined action preference vector $\boldsymbol{p'}$. This is exactly in accordance with the semantic relationship that the refine module should refine the action preference according to the state.
As shown in the dashed box in Figure~\ref{kgpn}, each hypernetwork takes the state $s$ as input and produces the weights and biases of one layer of $g(\cdot)$. Formally:
\begin{equation} \label{hyper_rm1}
\boldsymbol{p'} = g_{\theta}(\boldsymbol{p})
\end{equation}
where:
\begin{equation} \label{hyper_rm2}
\theta = h_{\alpha}(s) 
\end{equation}
Here $h_{\alpha}(\cdot)$ is the hypernetwork that takes state as input and output the weights $\theta$ for the refine module $g_{\theta}(\cdot)$. In both methods, a $sigmoid$ function is used finally to normalize the output of the refine module $g_{\theta}(\cdot)$. 

\begin{figure*}[t]
\centering
\subfigure[$CartPole$]{
\begin{minipage}[t]{0.24\textwidth}
\centering
\includegraphics[width=0.99\textwidth]{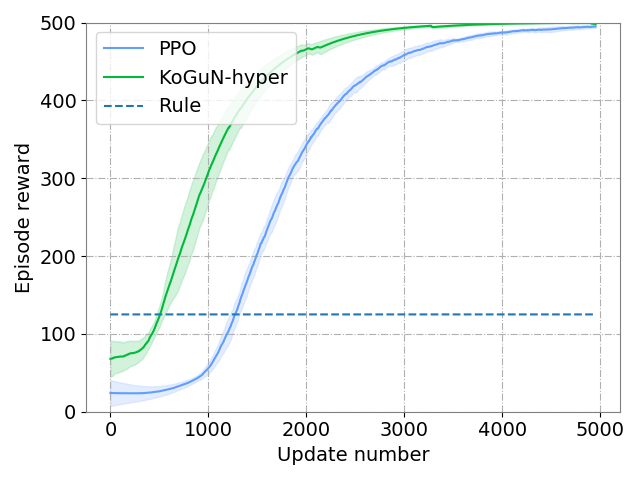}
% \caption{$CartPole$}
\label{cpresult}
\end{minipage}%
}%
\subfigure[$LunarLander$]{
\begin{minipage}[t]{0.24\textwidth}
\centering
\includegraphics[width=0.99\textwidth]{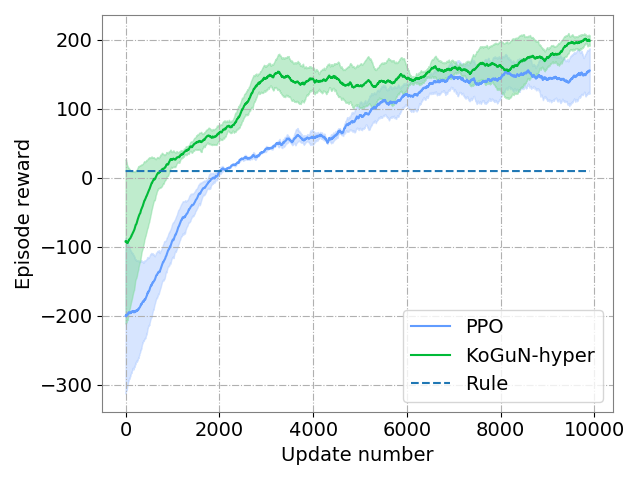}
% \caption{$LunarLander$}
\label{llresult}
\end{minipage}%
}%
\subfigure[$FlappyBird$]{
\begin{minipage}[t]{0.24\textwidth}
\centering
\includegraphics[width=0.99\textwidth]{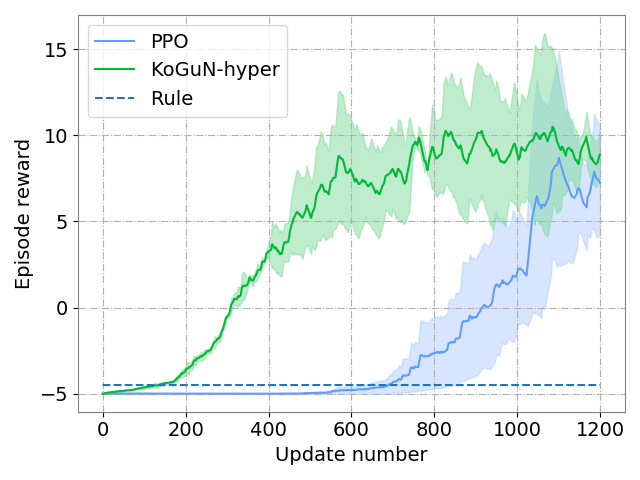}
% \caption{$FlappyBird$}
\label{fbresult}
\end{minipage}
}%
\subfigure[$LunarLanderContinuous$]{
\begin{minipage}[t]{0.24\textwidth}
\centering
\includegraphics[width=0.99\textwidth]{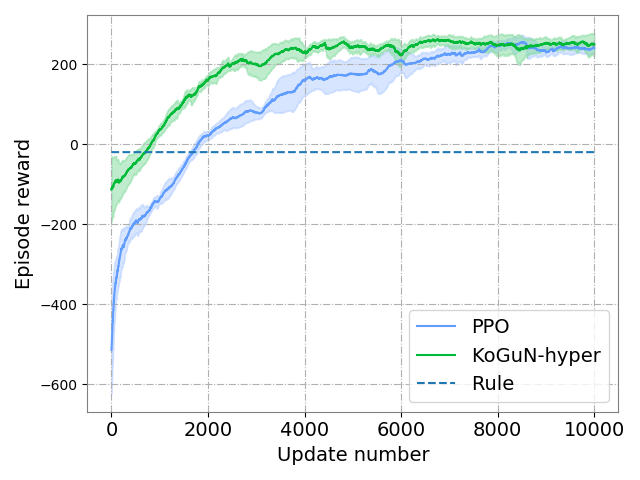}
% \caption{$ContinuousLunarLander$}
\label{cllresult}
\end{minipage}
}%
\centering
\caption{Experimental results for PPO without KoGuN ($blue$), KoGuN with hypernetworks for refine module ($green$) and pure knowledge controller ($dashed~line$) on four tasks. The shaded region denotes standard deviation of average evaluation over 5 trials. }
\label{allresult}
\end{figure*}

\begin{figure*}[t]
\centering
\includegraphics[width=0.9\textwidth]{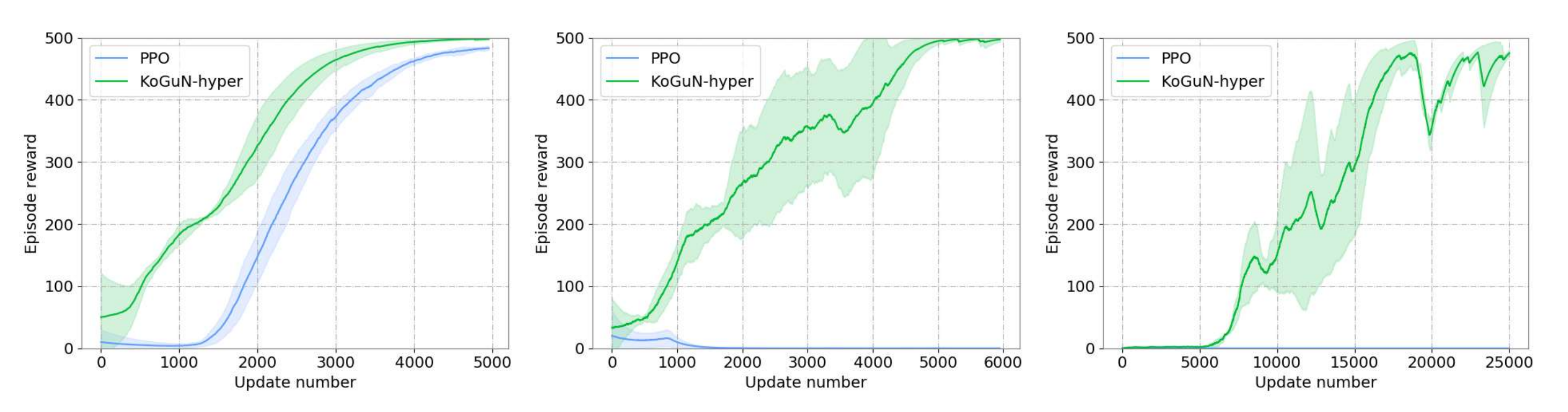} 
\caption{Experimental results in the setting of sparse reward. We set $d$=50 ($left$), 100 ($middle$) and 250 ($right$) in the task $CartPole$.}
\label{delay_result}
\end{figure*}

\begin{figure}[t]
    \centering
    \includegraphics[width=0.9\columnwidth]{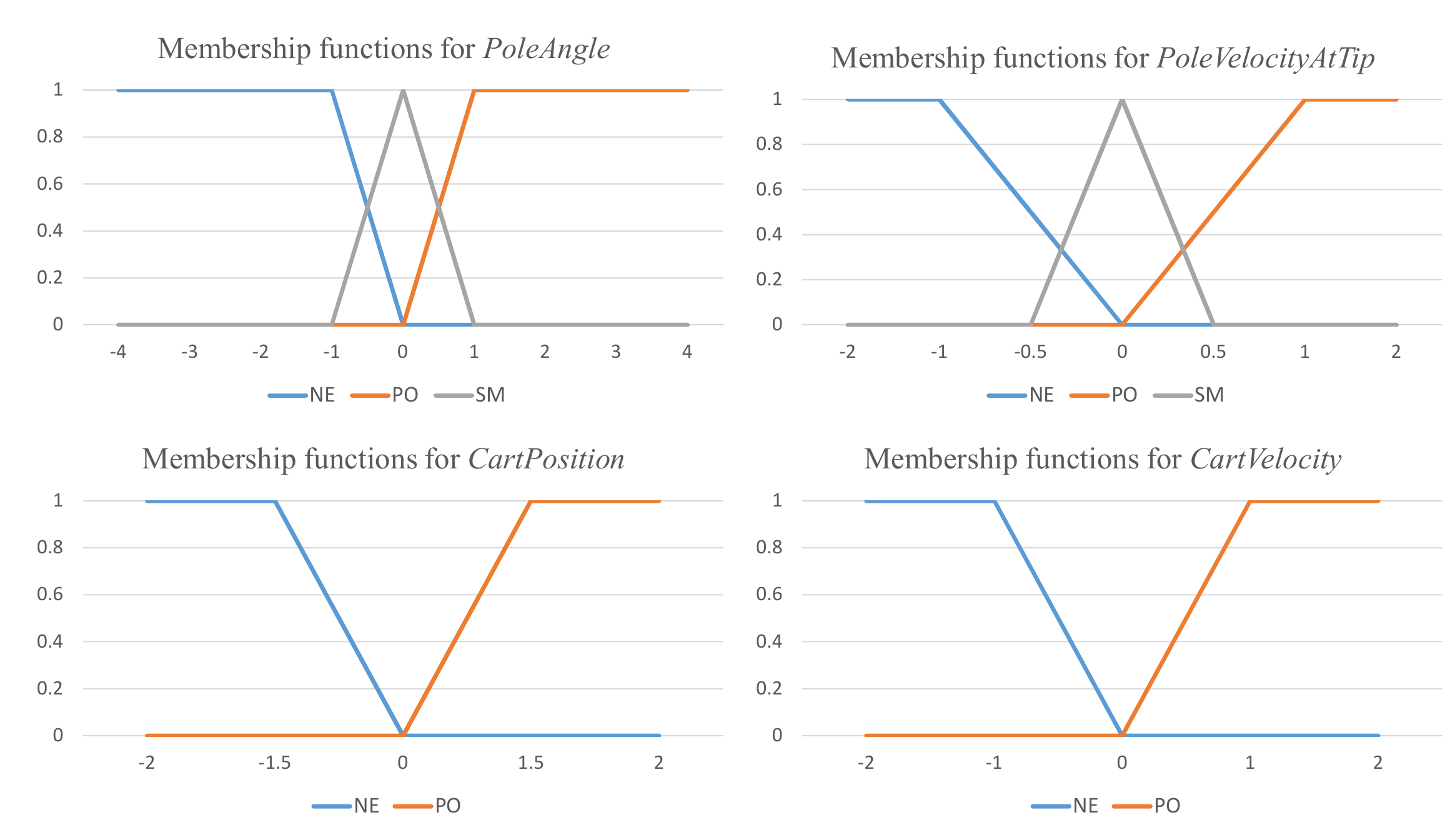} 
    \caption{Membership functions used in $CartPole$.}
    \label{cpmfs}
\end{figure}

\subsection{Integrating Knowledge Controller and Refine Module} \label{combine}
To leverage prior knowledge, we use the rough policy given by the knowledge controller to guide the agent to interact with environments at early stage. Therefore, as demonstrated in Figure~\ref{kgpn}, a weighted sum between $\boldsymbol{p}$ and $\boldsymbol{p'}$ is used as the final preference vector. At the early stage of the training phase, $\boldsymbol{p}$ provided by the knowledge controller, accounting for a larger proportion. As the training continues, the proportion of $\boldsymbol{p'}$ gradually increases. For tasks with discrete action space, applying $softmax$ to the weighted sum, we get the final output of KoGuN:
\begin{equation}
    a \sim softmax(\frac{w_{1}\boldsymbol{p}+w_{2}\boldsymbol{p'}}{\tau})
\end{equation}
where:
\begin{equation}
    w_1+w_2=1
\end{equation}
Here $\tau$ is the temperature to sharpen the action distribution. And for tasks with continuous action space, the mean of the action Gaussian distribution is the weighted sum of $\boldsymbol{p}$ and $\boldsymbol{p'}$:
\begin{equation}
    a\sim \mathcal{N}(w_1\boldsymbol{p}+w_2\boldsymbol{p'}, \sigma)
\end{equation}
The weight corresponding to the knowledge controller $w_1$ decays linearly to a small value as the training continues. 
At beginning, the refine module is not well trained so the gradient flow from the refine module may harm the learning of the knowledge controller.  To stabilize training process, we prohibit the gradient from the refine module back propagating to the knowledge controller and for forcing the refine module to adapt to the knowledge controller. $red ~dashed ~lines$ in Figure~\ref{kgpn} denote the backward gradient flow.

KoGuN is an end-to-end policy framework, thus it can be combined with any policy-based algorithm. In this paper, we use proximal policy optimization (PPO) \cite{ppo} as our basic RL algorithm.

\section{Experiments} \label{exp}

In this section, we firstly evaluate our algorithm on four tasks in Section~\ref{eval}
% (see Figure~\ref{envs})
: $Cart Pole$ \cite{cartpole}, $LunarLander$ and $LunarLanderContinuous$ in OpenAI Gym \cite{gym} and $FlappyBird$ in PLE \cite{ple}. Moreover, we show the effectiveness and robustness of KoGuN in sparse reward setting in Section~\ref{sparse}. Then we finally conduct ablation study to better understanding the contribution of each part of our framework in Section~\ref{ablation}. 

The experimental setup is as follows: for all the tasks, we use Adam optimizer
\cite{adam} 
with a learning rate of $1\times10^{-4}$ and the temperature $\tau=0.1$. The discounted factor $\gamma$ is set to 0.99 and the GAE $\lambda$ is set to 0.95. The policy is updated every 128 timesteps. For PPO without KoGuN, we use a neural network with two full-connected hidden layers as policy approximator. For KoGuN with normal network (KoGuN-concat) as refine module, we also use a neural network with two full-connected hidden layers for the refine module. For KoGuN with hypernetworks (KoGuN-hyper), we use hypernetworks to generate a refine module with one hidden layer. Each hypernetwork has two hidden layers. All hidden layers described above have 32 units. $w_{1}$ is set to 0.7 at beginning and decays to 0.1 in the end of training phase. 

\begin{figure*}[t]
\centering
\subfigure[$CartPole$]{
\begin{minipage}[t]{0.24\textwidth}
\centering
\includegraphics[width=0.99\textwidth]{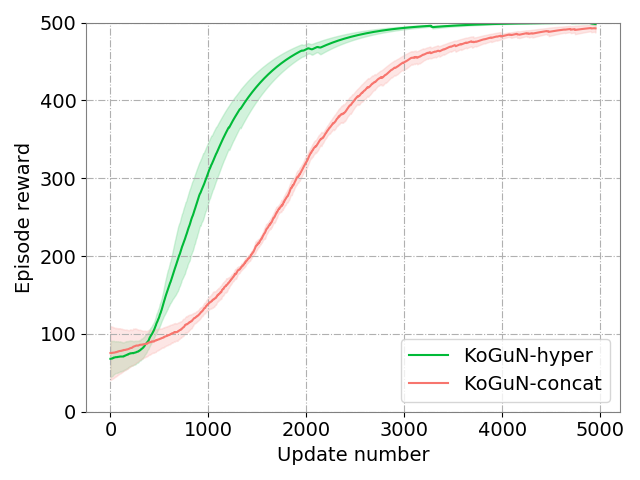}
% \caption{$CartPole$}
\label{cpab}
\end{minipage}%
}%
\subfigure[$LunarLander$]{
\begin{minipage}[t]{0.24\textwidth}
\centering
\includegraphics[width=0.99\textwidth]{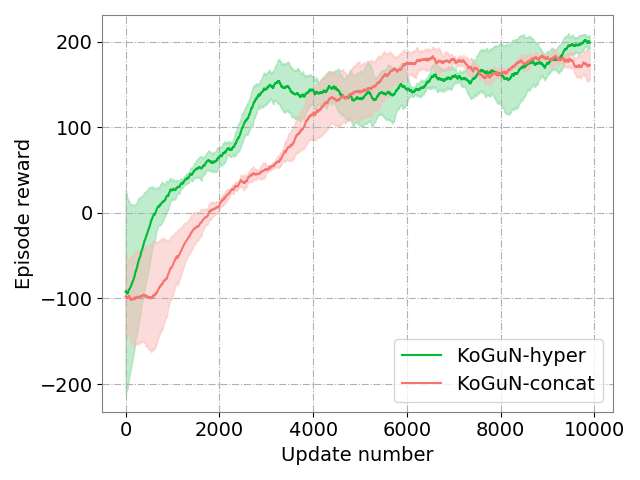}
% \caption{$LunarLander$}
\label{llab}
\end{minipage}%
}%
\subfigure[$FlappyBird$]{
\begin{minipage}[t]{0.24\textwidth}
\centering
\includegraphics[width=0.99\textwidth]{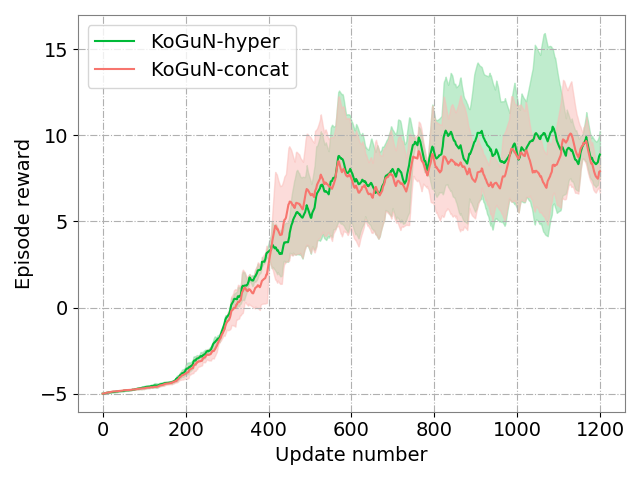}
% \caption{$FlappyBird$}
\label{fbab}
\end{minipage}
}%
\subfigure[$LunarLanderContinuous$]{
\begin{minipage}[t]{0.24\textwidth}
\centering
\includegraphics[width=0.99\textwidth]{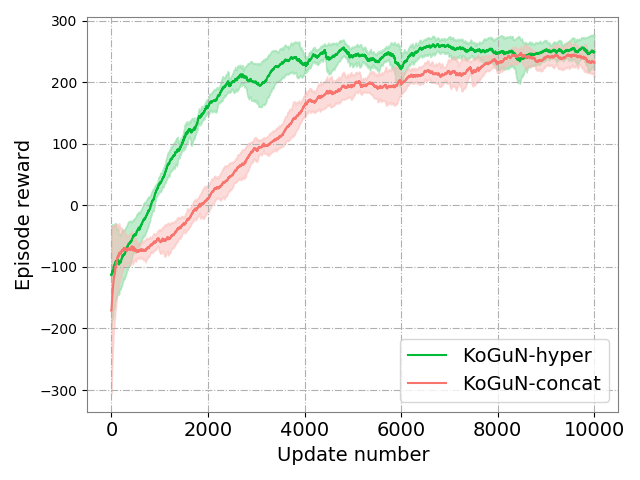}
% \caption{$ContinuousLunarLander$}
\label{cllab}
\end{minipage}
}%
\centering
\caption{Experimental results for KoGuN with hypernetworks for refine module (KoGuN-hyper) and KoGuN with normal neural networks for refine module (KoGuN-concat) on four tasks. }
\label{abresult}
\end{figure*}
\begin{figure}[t]
\centering
\subfigure[]{
\begin{minipage}[t]{0.21\textwidth}
\centering
\includegraphics[width=0.99\textwidth]{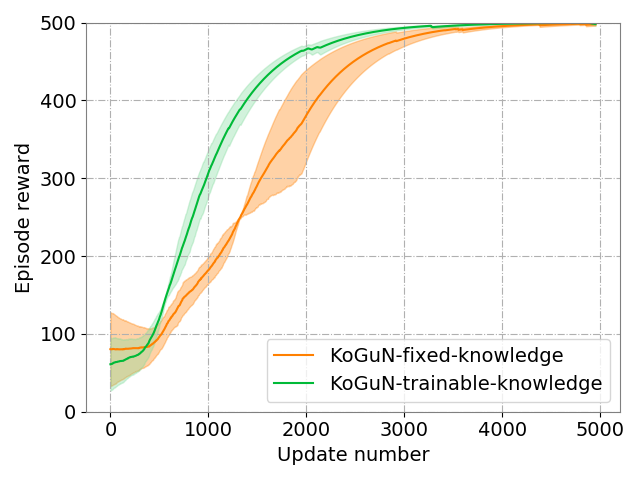}
\label{trainable_result}
\end{minipage}%
}%
\subfigure[]{
\begin{minipage}[t]{0.21\textwidth}
\centering
\includegraphics[width=0.99\textwidth]{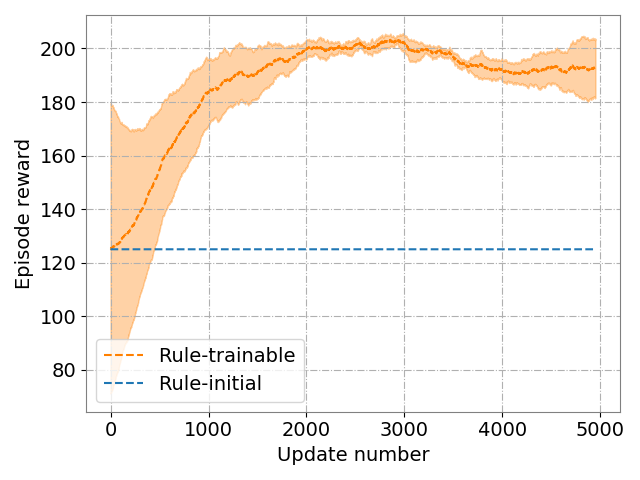}
\label{rule_p}
\end{minipage}%
}%
\centering
\caption{KoGuN with trainable knowledge controller $vs.$ KoGuN with fixed knowledge controller ($left$). Performance of knowledge controller during training ($right$).}
\end{figure}

\subsection{Evaluation} \label{eval}
We use the environments mentioned above as our evaluation environments. We designed some rules for each task. Because of space limitation, here we only give the rules used in $CartPole$ as an example. In $CartPole$, a pole is attached by an un-actuated joint to a cart. The system is controlled by applying a force to the cart. The episode ends when the pole is more than 15 degrees from vertical, or the cart moves more than 2.4 units away from the center. The state is represented by a four-dimensional vector $[CartPosition, CartVelocity,$ $ PoleAngle, PoleVelocityAtTip]$, which ranges from $[-2.4, -Inf, -15, -Inf]$ to $[2.4, Inf, 15, Inf]$. There are two available actions $[p, n]$: push the cart towards positive direction or negative direction. Totally 6 rules are used in the experiment and they are listed below:
\begin{itemize}
    \item $Rule$ 1: IF $PoleAngle$ is $NE$ and $PoleVelocityAtTip$ is $NE$ THEN $Action$ is $n$.
    \item $Rule$ 2: IF $PoleAngle$ is $PO$ and $PoleVelocityAtTip$ is $PO$ THEN $Action$ is $p$.
    \item $Rule$ 3: IF $PoleAngle$ is $SM$ and $PoleVelocityAtTip$ is $NE$ THEN $Action$ is $n$.
    \item $Rule$ 4: IF $PoleAngle$ is $SM$ and $PoleVelocityAtTip$ is $PO$ THEN $Action$ is $p$.
    \item $Rule$ 5: IF $PoleAngle$ is $SM$ and $PoleVelocityAtTip$ is $SM$ and $CartPosition$ is $NE$ and $CartVelocity$ is $NE$ THEN $Action$ is $p$.
    \item $Rule$ 6: IF $PoleAngle$ is $SM$ and $PoleVelocityAtTip$ is $SM$ and $CartPosition$ is $PO$ and $CartVelocity$ is $PO$ THEN $Action$ is $n$.
\end{itemize}
$NE$, $PO$ and $SM$ are fuzzy sets which means $negative$, $positive$ and $small$ respectively. Their membership function are shown in Figure~\ref{cpmfs}. These rules are not well-crafted and are easy to understand. $Rule$ 1-$Rule$ 4 are designed to maintain the balance of the pole. 
$Rule$ 5 and $Rule$ 6 are designed to prevent the cart from moving out of the restricted range.

Figure~\ref{allresult} shows the experimental results on the four tasks. We can see that KoGuN can converge with much less update number. We also plot mean performance of pure knowledge controller (dash line in Figure~\ref{allresult}). We can see that KoGuN can dramatically accelerate the learning process even with such poor performance rule knowledge, both in discrete control and continuous control domain.

\subsection{Sparse Reward Setting} \label{sparse}
We further demonstrate the effectiveness of KoGuN in the setting of sparse reward in this section. We consider a sparse reward setting: multi-step accumulated rewards are given at sparse time steps. To simulate this setting, we deliver $d$-step accumulated reward every $d$ time steps in $CartPole$. 
We evaluate our algorithm under different delayed steps. Figure~\ref{delay_result} plots the results under sparse reward setting.

KoGuN shows superior performance in sparse reward setting. As the increase of delay step $d$, PPO without KoGuN can hardly learn a effective policies because the chance for the agent to get a reward signal is quite small. In contrast, with human knowledge, KoGuN enables the agent to avoid unnecessary exploration. As a consequence, the probability that the agent can obtain reward signals is greatly improved. Therefore, agents with KoGuN can learn effective policies even with very sparse reward signals.

\subsection{Ablation Study} \label{ablation}
\textbf{Hypernetworks as refine module:} To compare the performance of the two types of refine module described in Section~\ref{refine}, we conduct ablation experiments on all the four tasks. The experiment results are shown in Figure~\ref{abresult}. As described in Section~\ref{refine}, hypernetworks can more easily capture the complex relationship between the action preference vector $\boldsymbol{p}$ and the refined action preference vector $\boldsymbol{p'}$ in different states than normal neural networks. As a result, KoGuN with hypernetworks shows overall better performance than KoGuN with normal networks.

\noindent\textbf{Trainable knowledge controller:} We conduct another group of experiments on $CartPole$ to demonstrate the benefits of trainable knowledge controller. We compare trainable knowledge controller and fixed knowledge controller in Figure~\ref{trainable_result}. The performance of trainable controller during training is given in Figure~\ref{rule_p}, which means that the prior knowledge is constantly optimized during training. Trainable controller enables the provided knowledge to adapt as much as possible to the current task. As a result, agent can make better use of these knowledge and learns faster.

\section{Conclusion} \label{conc}
In this paper, we propose a novel policy network framework called KoGuN to leverage human knowledge to accelerate the learning process of RL agents. The proposed framework consists of a knowledge controller and a refine module. The knowledge controller represents human suboptimal knowledge using fuzzy rules and the refine module refines the imprecise prior knowledge. The policy framework is end-to-end and can be combined with existing policy-based algorithms to deal with tasks with both discrete action space and continuous action space. We evaluate our method on both discrete and continuous tasks and the experimental results show our method can significantly improve the learning efficiency of RL agents even with very low-performance human prior knowledge. Besides, the empirical results on sparse reward setting also demonstrate that the proposed method can alleviate the problem of sparse reward signals. As future work, we would like to investigate the knowledge representation method of more challenging tasks, such as tasks with high-dimensional visual data as state space. 
\bibliographystyle{named}
\bibliography{ijcai20}

\begin{thebibliography}{}

\bibitem[\protect\citeauthoryear{Barto and Sutton}{1982}]{cartpole}
Andrew~G Barto and Richard~S Sutton.
\newblock Simulation of anticipatory responses in classical conditioning by a
  neuron-like adaptive element.
\newblock {\em Behavioural Brain Research}, 4(3):221--235, 1982.

\bibitem[\protect\citeauthoryear{Berenji}{1992}]{rlfuzzy}
Hamid~R Berenji.
\newblock A reinforcement learning-based architecture for fuzzy logic control.
\newblock {\em International Journal of Approximate Reasoning}, 6(2):267--292,
  1992.

\bibitem[\protect\citeauthoryear{Brockman \bgroup \em et al.\egroup
  }{2016}]{gym}
Greg Brockman, Vicki Cheung, Ludwig Pettersson, Jonas Schneider, John Schulman,
  Jie Tang, and Wojciech Zaremba.
\newblock Openai gym, 2016.

\bibitem[\protect\citeauthoryear{Collobert \bgroup \em et al.\egroup
  }{2011}]{augfeat}
Ronan Collobert, Jason Weston, L{\'e}on Bottou, Michael Karlen, Koray
  Kavukcuoglu, and Pavel Kuksa.
\newblock Natural language processing (almost) from scratch.
\newblock {\em Journal of machine learning research}, 12(Aug):2493--2537, 2011.

\bibitem[\protect\citeauthoryear{Fischer \bgroup \em et al.\egroup
  }{2019}]{dl2}
Marc Fischer, Mislav Balunovic, Dana Drachsler-Cohen, Timon Gehr, Ce~Zhang, and
  Martin Vechev.
\newblock Dl2: Training and querying neural networks with logic.
\newblock In {\em International Conference on Machine Learning}, pages
  1931--1941, 2019.

\bibitem[\protect\citeauthoryear{Garcez \bgroup \em et al.\egroup }{2012}]{nss}
Artur S~d'Avila Garcez, Krysia~B Broda, and Dov~M Gabbay.
\newblock {\em Neural-symbolic learning systems: foundations and applications}.
\newblock Springer Science \& Business Media, 2012.

\bibitem[\protect\citeauthoryear{Gordon \bgroup \em et al.\egroup
  }{2011}]{dagger}
Geoffrey~J. Gordon, David~B. Dunson, and Miroslav Dud{\'{\i}}k, editors.
\newblock {\em Proceedings of the Fourteenth International Conference on
  Artificial Intelligence and Statistics, {AISTATS} 2011, Fort Lauderdale, USA,
  April 11-13, 2011}, volume~15 of {\em {JMLR} Proceedings}. JMLR.org, 2011.

\bibitem[\protect\citeauthoryear{Ha \bgroup \em et al.\egroup
  }{2016}]{hypernet}
David Ha, Andrew Dai, and Quoc~V Le.
\newblock Hypernetworks.
\newblock {\em arXiv preprint arXiv:1609.09106}, 2016.

\bibitem[\protect\citeauthoryear{Hester \bgroup \em et al.\egroup
  }{2017}]{dqfd}
Todd Hester, Matej Vecer{\'{\i}}k, Olivier Pietquin, Marc Lanctot, Tom Schaul,
  Bilal Piot, Andrew Sendonaris, Gabriel Dulac{-}Arnold, Ian Osband, John
  Agapiou, Joel~Z. Leibo, and Audrunas Gruslys.
\newblock Learning from demonstrations for real world reinforcement learning.
\newblock {\em CoRR}, abs/1704.03732, 2017.

\bibitem[\protect\citeauthoryear{Ho and Ermon}{2016}]{gail}
Jonathan Ho and Stefano Ermon.
\newblock Generative adversarial imitation learning.
\newblock In {\em Advances in neural information processing systems}, pages
  4565--4573, 2016.

\bibitem[\protect\citeauthoryear{Hu \bgroup \em et al.\egroup
  }{2016}]{harnessing}
Zhiting Hu, Xuezhe Ma, Zhengzhong Liu, Eduard Hovy, and Eric Xing.
\newblock Harnessing deep neural networks with logic rules.
\newblock {\em arXiv preprint arXiv:1603.06318}, 2016.

\bibitem[\protect\citeauthoryear{Kingma and Ba}{2015}]{adam}
Diederik~P. Kingma and Jimmy Ba.
\newblock Adam: {A} method for stochastic optimization.
\newblock In {\em 3rd International Conference on Learning Representations,
  {ICLR} 2015, San Diego, CA, USA, May 7-9, 2015, Conference Track
  Proceedings}, 2015.

\bibitem[\protect\citeauthoryear{Mnih \bgroup \em et al.\egroup }{2015}]{dqn}
V~Mnih, K~Kavukcuoglu, D~Silver, A.~A. Rusu, J~Veness, M.~G. Bellemare,
  A~Graves, M~Riedmiller, A.~K. Fidjeland, and G~Ostrovski.
\newblock Human-level control through deep reinforcement learning.
\newblock {\em Nature}, 518(7540):529, 2015.

\bibitem[\protect\citeauthoryear{Pomerleau}{1991}]{bc}
Dean Pomerleau.
\newblock Efficient training of artificial neural networks for autonomous
  navigation.
\newblock {\em Neural Computation}, 3(1):88--97, 1991.

\bibitem[\protect\citeauthoryear{Schmidhuber}{1992}]{fastweights}
J{\"u}rgen Schmidhuber.
\newblock Learning to control fast-weight memories: An alternative to dynamic
  recurrent networks.
\newblock {\em Neural Computation}, 4(1):131--139, 1992.

\bibitem[\protect\citeauthoryear{Schulman \bgroup \em et al.\egroup
  }{2015a}]{robotic}
John Schulman, Sergey Levine, Pieter Abbeel, Michael Jordan, and Philipp
  Moritz.
\newblock Trust region policy optimization.
\newblock In {\em International conference on machine learning}, pages
  1889--1897, 2015.

\bibitem[\protect\citeauthoryear{Schulman \bgroup \em et al.\egroup
  }{2015b}]{gae}
John Schulman, Philipp Moritz, Sergey Levine, Michael Jordan, and Pieter
  Abbeel.
\newblock High-dimensional continuous control using generalized advantage
  estimation.
\newblock {\em arXiv preprint arXiv:1506.02438}, 2015.

\bibitem[\protect\citeauthoryear{Schulman \bgroup \em et al.\egroup
  }{2017}]{ppo}
John Schulman, Filip Wolski, Prafulla Dhariwal, Alec Radford, and Oleg Klimov.
\newblock Proximal policy optimization algorithms.
\newblock {\em arXiv preprint arXiv:1707.06347}, 2017.

\bibitem[\protect\citeauthoryear{Silver \bgroup \em et al.\egroup
  }{2016}]{mastergo}
D~Silver, A.~Huang, C.~J. Maddison, A~Guez, L~Sifre, den Driessche~G Van,
  J~Schrittwieser, I~Antonoglou, V~Panneershelvam, and M~Lanctot.
\newblock Mastering the game of go with deep neural networks and tree search.
\newblock {\em Nature}, 529(7587):484--489, 2016.

\bibitem[\protect\citeauthoryear{Sutton and Barto}{2018}]{suttonbook}
Richard~S Sutton and Andrew~G Barto.
\newblock {\em Reinforcement learning: An introduction}.
\newblock MIT press, 2018.

\bibitem[\protect\citeauthoryear{Tasfi}{2016}]{ple}
Norman Tasfi.
\newblock Pygame learning environment.
\newblock \url{https://github.com/ntasfi/PyGame-Learning-Environment}, 2016.

\bibitem[\protect\citeauthoryear{Wu \bgroup \em et al.\egroup }{2017}]{acktr}
Yuhuai Wu, Elman Mansimov, Roger~B Grosse, Shun Liao, and Jimmy Ba.
\newblock Scalable trust-region method for deep reinforcement learning using
  kronecker-factored approximation.
\newblock In {\em Advances in neural information processing systems}, pages
  5279--5288, 2017.

\bibitem[\protect\citeauthoryear{Yager and Zadeh}{2012}]{matchmethod}
Ronald~R Yager and Lotfi~A Zadeh.
\newblock {\em An introduction to fuzzy logic applications in intelligent
  systems}, volume 165.
\newblock Springer Science \& Business Media, 2012.

\bibitem[\protect\citeauthoryear{Zadeh}{1965}]{fuzzysets}
L.~A. Zadeh.
\newblock Fuzzy sets.
\newblock {\em Information and Control}, 8(3):338--353, 1965.

\bibitem[\protect\citeauthoryear{Zadeh}{1996}]{krfl}
Lotfi~A Zadeh.
\newblock Knowledge representation in fuzzy logic.
\newblock In {\em Fuzzy Sets, Fuzzy Logic, And Fuzzy Systems: Selected Papers
  by Lotfi A Zadeh}, pages 764--774. World Scientific, 1996.

\end{thebibliography}
\end{document}